\documentclass[runningheads,a4paper]{llncs}

\usepackage{amssymb}
\usepackage{graphicx}

\usepackage{url}
\urldef{\mailsa}\path|{alfred.hofmann, ursula.barth, ingrid.haas, frank.holzwarth,|
\urldef{\mailsb}\path|anna.kramer, leonie.kunz, christine.reiss, nicole.sator,|
\urldef{\mailsc}\path|erika.siebert-cole, peter.strasser, lncs}@springer.com|    
\newcommand{\keywords}[1]{\par\addvspace\baselineskip
\noindent\keywordname\enspace\ignorespaces#1}

\usepackage{algorithm}
\usepackage{algorithmic}

\usepackage{hyperref}

\usepackage{xspace}
\usepackage{amsfonts,amstext,amsmath,amssymb}
\usepackage{multirow}
\usepackage{array,url}
\usepackage{color}

\newcommand{\kb}[1]{{\small\texttt{#1}}\xspace}
\newcommand{\txt}[1]{\textit{#1}}

\renewcommand{\Re}{\mathbb{R}}
\newcommand{\bbw}{\boldsymbol{w}}
\newcommand{\bbW}{\boldsymbol{W}}
\newcommand{\bbX}{\boldsymbol{X}}
\newcommand{\bbM}{\boldsymbol{M}}
\newcommand{\bbv}{\boldsymbol{v}}
\newcommand{\bbV}{\boldsymbol{V}}

\newcommand{\bbg}{\boldsymbol{g}}
\newcommand{\bbf}{\boldsymbol{f}}

\newcommand{\norm}[1]{\parallel\! #1\! \parallel}

\newcommand{\Sft}[2]{S_{{\rm ft}}(#1,#2)}
\newcommand{\Sftv}{S_{{\rm ft}}}
\newcommand{\Spr}[2]{S_{{\rm prp}}(#1,#2)}
\newcommand{\Sprv}{S_{{\rm prp}}}

\newcommand{\card}[1]{|#1|}

\newcommand{\ab}[1]{{\color{black}{#1}}}

\newcommand{\dbp}{{\sc DBpedia}\xspace}
\newcommand{\fb}{{\sc Freebase}\xspace}
\newcommand{\rv}{{\sc ReVerb}\xspace}
\newcommand{\wk}{{\sc WikiAnswers}\xspace}
\newcommand{\wq}{{\sc WebQuestions}\xspace}
\newcommand{\wkrv}{{\sc WikiAnswers+ReVerb}\xspace}

\begin{document}

\mainmatter  

\title{Open Question Answering  with Weakly Supervised Embedding Models}

\titlerunning{Open Question Answering  with Weakly Supervised Embedding Models}

%
%
\author{Antoine Bordes$^\dagger$
\and Jason Weston$^\ddagger$ \and Nicolas Usunier$^\dagger$}
%

\institute{$^\dagger$    Universit\'e de Technologie de Compi\`egne -- CNRS,\\
Heudiasyc UMR 7253, Compi\`egne, France \\
$^\ddagger$ Google, 111 8th avenue, New York, NY, USA\\
\path|{fbordesan, nusunier@utc.fr, jaseweston@gmail.com}|
}

%
%

\toctitle{Lecture Notes in Computer Science}
\tocauthor{Authors' Instructions}
\maketitle

\begin{abstract}
Building computers able to answer questions on any subject is a long standing goal of artificial intelligence.
%
Promising progress has recently been achieved by 
%
methods that learn to map questions to logical forms or database queries.
Such approaches can be effective but at the cost of 
either large amounts of human-labeled data or by 
defining lexicons and grammars tailored by practitioners.
%
In this paper, we instead take the radical approach of learning to map questions to vectorial feature representations. 
By mapping answers into the same space one can query any knowledge base 
independent of its schema, without requiring any grammar or lexicon.
Our method is trained with a new optimization procedure combining stochastic gradient descent followed by a fine-tuning step using the weak supervision provided by blending automatically and collaboratively generated resources.
%
%
We empirically demonstrate that our model can capture meaningful signals from its noisy supervision leading to major improvements over {\sc paralex}, the only existing method able to be trained 
on similar weakly labeled data.
\keywords{natural language processing, question answering, weak supervision, embedding models}
\end{abstract}

\section{Introduction}
\label{sec:intro}

This paper addresses the challenging problem of open-domain question answering, which consists of building systems able to answer questions from any domain.
Any advance on this difficult topic would bring a huge leap forward in building new ways of accessing knowledge.
An important development in this area has been the creation of large-scale Knowledge Bases (KBs), such as \fb \cite{bollacker2008freebase} and \dbp \cite{dbpedia-swj} which store huge amounts of general-purpose information.
They are organized as databases of triples connecting pairs of entities by various relationships and of the form (\kb{left entity}, \kb{relationship}, \kb{right entity}). 
Question answering is then defined as the task of retrieving the correct entity or set of entities from a KB given a query expressed as a question in natural language.

The use of KBs simplifies the problem by separating the issue of collecting and organizing information (i.e. {\it information extraction}) from the one of searching through it (i.e. {\it question answering } or {\it natural language interfacing}).
However, open question answering remains challenging because of the scale of these KBs (billions of triples, millions of entities and relationships) and of the difficulty for machines to interpret natural language.
Recent progress \cite{cai-yates:2013:ACL2013,berant-EtAl:2013:EMNLP,kwiatkowski-EtAl:2013:EMNLP,paralex} has been made by tackling this problem with semantic parsers. These methods convert questions into logical forms or database queries (e.g. in SPARQL) which are then subsequently used to query KBs for answers.
Even if such systems have shown the ability to handle large-scale KBs, they require practitioners to hand-craft lexicons, grammars, and KB schema for the parsing to be effective. This non-negligible human intervention might not be generic enough to conveniently scale up to new databases with other schema, broader vocabularies or other languages than English.

In this paper, we instead take the approach of converting questions to (uninterpretable) vectorial representations which require no pre-defined grammars or lexicons and can query any KB independent of its schema.
Following \cite{paralex}, we focus on answering simple factual questions on a broad range of topics, more specifically, those for which single KB triples stand for both the question and an answer (of which there may be many). For example, (\kb{parrotfish.e}, \kb{live-in.r}, \kb{southern-water.e}) stands for {\it What is parrotfish's habitat?} and \kb{southern-water.e} and (\kb{cantonese.e}, \kb{be-major-language-in.r}, \kb{hong-kong.e}) for {\it What is the main language of Hong-Kong?} and \kb{cantonese.e}.
In this task, the main difficulties come from lexical variability rather than from complex syntax, having multiple answers per question, and the absence of a supervised training signal.

Our approach is based on learning low-dimensional vector embeddings of words and of KB triples so that representations of questions and corresponding answers end up being similar in the embedding space.
Unfortunately, we do not have access to any human labeled (query, answer) supervision for this task.
In order to avoid transferring the cost of manual intervention to the one of labeling large amounts of data,  we make use of weak supervision.
We show empirically that our model is able to take advantage of noisy and indirect supervision by (i) automatically generating questions from KB triples and treating this as training data; and (ii) supplementing this with a data set of questions collaboratively marked as paraphrases but with no associated answers.
We end up learning meaningful vectorial representations for questions involving up to 800k words and for triples of an mostly automatically created KB with 2.4M entities and 600k relationships. Our method strongly outperforms previous results on the \wkrv evaluation data set introduced by \cite{paralex}.
Even if the embeddings obtained after training are of good quality, the scale of the optimization problem makes it hard to control and to lead to convergence. 
Thus, we propose a method to fine-tune embedding-based models by carefully optimizing a matrix parameterizing the similarity used in the embedding space, leading to  a consistent improvement in performance.

The rest of the  paper is organized as follows. Section~\ref{sec:rwork} discusses some previous work and Section~\ref{sec:qa} introduces the problem of open question answering. Then, Section~\ref{sec:model} presents our model and Section~\ref{sec:exp} our experimental results.

\section{Related Work}
\label{sec:rwork}

Large-scale question answering has a long history, mostly initiated via the TREC tracks~\cite{voorhees2000building}.
The first successful systems transformed the questions into queries which were fed to web search engines, the answer being subsequently extracted from top returned pages or snippets \cite{kwok2001scaling,banko2002askmsr}.
Such approaches require significant engineering to hand-craft queries and then parse and search over results.

The emergence of large-scale KBs, such as \fb \cite{bollacker2008freebase} or \dbp \cite{dbpedia-swj}, changed the setting by transforming open question answering into a problem of querying a KB using natural language.
This is a challenging problem, which would  require huge amount of labeled data to be tackled properly by purely supervised machine learning methods because of the great variability of language and of the large scale of KBs.
The earliest methods for open question-answering with KBs, based on hand-written templates \cite{yahya2012natural,unger2012template}, were not robust enough to such variability over possibly evolving KBs (addition/deletion of triples and entities).
%
The solution to gain more expressiveness via machine learning comes from distant or indirect supervision to circumvent the issue of labeled data.
Initial works attempting to learn to connect KBs and natural language with less supervision have actually been tackling the information extraction problem \cite{mintz2009distant,hoffmann2011knowledge,lao2012reading,riedel2013relation}.

Recently, new systems for learning question answering systems with few labeled data have been introduced
based on semantic parsers \cite{cai-yates:2013:ACL2013,berant-EtAl:2013:EMNLP,kwiatkowski-EtAl:2013:EMNLP}. Such works tend to require realistic amounts of manual intervention via labeled examples, but still need vast efforts to carefully design lexicons, grammars and the KB.
In contrast, \cite{paralex} proposed a framework for open question answering requiring little human annotation. 
Their system, {\sc Paralex}, answers questions with more limited semantics than those introduced in \cite{berant-EtAl:2013:EMNLP,kwiatkowski-EtAl:2013:EMNLP}, but does so at a very large scale in an open-domain manner. It is trained using automatically and collaboratively generated data and using the KB \rv \cite{ReVerb2011}.
In this work, we follow this trend by proposing an embedding-based model for question answering that is 
also trained under weak and cheap supervision.


Embedding-based models are getting more and more popular in natural language processing.
Starting from the neural network language model of \cite{bengio03}, these methods have now reached near state-of-the-art performance on many standard tasks while usually requiring less hand-crafted features \cite{collobert:2011b,socher2013recursive}.
Recently, some embedding models have been proposed to perform a connection between natural language and KBs for word-sense disambiguation \cite{bordes:12aistats} and for information extraction \cite{weston-EtAl:2013:EMNLP}. 
Our work builds on these approaches to instead learn to perform open question answering under weak supervision, which to our knowledge has not been attempted before.

\section{Open-domain Question Answering}
\label{sec:qa}

In this paper, we follow the question answering framework of \cite{paralex} and use the same data. Hence, relatively little labeling or feature engineering has been used.

\subsection{Task Definition}

Our work considers the task of question answering as in \cite{paralex}: given a question $q$, the corresponding answer is given by a triple $t$ from a KB.
\ab{This means that we consider questions for which a set of triples $t$ provide an interpretation of the question and its answer, such as:}\\
%
~\\
%
$\mbox{~~}\bullet\mbox{}q$: {\it What environment does a dodo live in ?}\\
$\mbox{~~~~~~}t$: \kb{(dodo.e, live-in.r, makassar.e)}\\
~\\
$\mbox{~~}\bullet\mbox{}q$: {\it What are the symbols for Hannukah ?}\\
$\mbox{~~~~~~}t$: \kb{(menorah.e, be-for.r, hannukah.e)}\\
~\\
$\mbox{~~}\bullet\mbox{}q$: {\it What is a laser used for?}\\
$\mbox{~~~~~~}t$: \kb{(hologram.e,be-produce-with.r,laser.e)}\\

Here, we only give a single $t$ per question, but many can exist.
In the remainder, the KB is denoted ${\cal K}$ and its set of entities and relationships is ${\cal E}$.
The word vocabulary for questions is termed ${\cal V}$.
$n_v$ and $n_e$ are the sizes of ${\cal V}$ and ${\cal E}$ respectively. 

Our model consists in learning a function $S(\cdot)$, which can score question-answer triple pairs $(q,t)$. Hence, finding the top-ranked answer $\hat{t}(q)$ to  a question $q$ is directly carried out by:
\[
    \hat{t}(q)=\arg \max_{t'\in{\cal K}} S(q,t')~.
\]
To handle multiple answer, we instead present the results as a ranked list, rather than taking the top prediction, and evaluate that instead. 

Using the scoring function $S(\cdot)$ allows to directly query the KB without needing to define an intermediate structured logical representation for questions as in semantic parsing systems. 
We aim at learning $S(\cdot)$, with no human-labeled supervised 
data in the form (question, answer) pairs,
but only by indirect supervision,
generated either automatically or collaboratively. 
We detail in the rest of this section our process for creating training data.

\subsection{Training Data}

Our training data consists of two sources: an automatically created KB, \rv,  from which we generate questions and a set of pairs of questions collaboratively labeled as paraphrases from the website \wk.

\paragraph{Knowledge Base}
The set of potential answers ${\cal K}$ is given by the KB \rv \cite{ReVerb2011}.
\rv is an open-source database composed of more than 14M triples, made of more 
than 2M entities and 600k relationships, which have been automatically extracted from 
the ClueWeb09 corpus \cite{pomikalek2012building}. In the following, entities are denoted with a \kb{.e} suffix and relationships with a \kb{.r} suffix. 

\rv contains broad and general knowledge harvested with very little human intervention, which suits the realistically supervised setting. 
But, as a result, \rv is ambiguous and noisy with many useless triples and entities as well as numerous duplicates. For instance, \kb{winston-churchill.e}, \kb{churchill.e} and even \kb{roosevelt-and-churchill.e} are all distinct entities.
Table~\ref{tab:exrv} presents some examples of triples: some make sense, some others are completely unclear or useless. 

In contrast to highly curated databases such \fb, \rv has more noise but also many more relation types (\fb has around 20k). So for some types of triple it has much better coverage, despite the larger size of \fb; for example \fb does not cover verbs like afraid-of or suffer-from.

\begin{table}[t]
\label{tab:exrv}
\begin{center}
\caption{Examples of triples from the KB \rv.} 
\begin{small}
\begin{tabular}{|l|}
\hline
\multicolumn{1}{|l|}{ {\bf \kb{left entity, relationship, right entity}  }}\\
\hline
\kb{churchill.e, be-man-of.r, great-accomplishment.e}\\
\kb{churchill-and-roosevelt.e, meet-in.r, cairo.e}\\
\kb{churchill.e, reply-on.r, may-19.e}\\
\kb{crick.e, protest-to.r, churchill.e}\\
\kb{churchill.e, leave-room-for.r, moment.e}\\
\kb{winston-churchill.e, suffer-from.r, depression.e}\\
\kb{churchill.e, be-prime-minister-of.r, great-britain.e}\\
\kb{churchill.e, die-in.r, winter-park.e}\\
\kb{winston-churchill.e, quote-on.r, mug.e}\\
\kb{churchill.e, have-only.r, compliment.e}\\
\hline
\end{tabular}
\end{small}
\end{center}
\end{table}

\paragraph{Questions Generation}

We have no available data of questions $q$ labeled with their answers, i.e. with the corresponding triples $t\in{\cal K}$. Following \cite{paralex}, we hence decided to create such question-triple pairs automatically.
These pairs are generated  using the 16 seed questions displayed in Table~\ref{tab:qpat}. 
At each round, we pick a triple at random and then generate randomly one of the seed questions. Note only triples with a \kb{*-in.r} relation (denoted  \kb{r-in} in Table~\ref{tab:qpat}) can generate from the pattern {\it where did \kb{e} \kb{r} ?}, for example, and similar for other constraints. Otherwise, the pattern is chosen randomly.  
Except for these exceptions, we used all 16 seed questions for all triples hence generating approximately $16 \times 14\rm{M}$ questions stored in a training set we denote ${\cal D}$.

The generated questions are imperfect and noisy and create a weak training signal. 
Firstly, their syntactic structure is rather simplistic, and real questions as posed by humans (such as in our actual test) can look quite different to them.
Secondly,  many generated questions do not correspond to semantically valid English sentences.
For instance, since the type of entities in \rv is unknown, a pattern like {\it who does \kb{e} \kb{r} ?} can be chosen for a triple where the type of “?” in \kb{(?, r, e)} is not a person, and similar for other types (e.g. {\it when}). 
Besides, for the strings representing entities and relationships in the questions, we simply used their names in \rv, replacing \kb{-} by spaces and stripping off their suffixes, i.e. the string representing \kb{winston-churchill.e} is simply {\it winston churchill}.
While this is often fine, this is also very limited and caused many incoherences in the data.
Generating questions with a richer KB than \rv, such as \fb or \dbp, would lead to better quality because typing and better lexicons could be used. However, this would contradict one of our motivations which is to train a system with as little human intervention as possible (and hence choosing \rv over hand-curated KBs).
%

\begin{table}[t]
\caption{Patterns for generating questions from \rv triples following \cite{paralex}.}
\label{tab:qpat}
\vskip -0.1in
\begin{center}
\begin{small}
\begin{tabular}{|l|l|}
\hline
\multicolumn{1}{|c|}{{\bf KB Triple}}  & \multicolumn{1}{c|}{{\bf Question Pattern}}\\
\hline
\kb{(?, r, e)} & {\it who \kb{r} \kb{e} ?}\\
\kb{(?, r, e)} & {\it what \kb{r} \kb{e} ?}\\
\kb{(e, r, ?)} & {\it who does \kb{e} \kb{r} ?}\\
\kb{(e, r, ?)} & {\it what does \kb{e} \kb{r} ?}\\
\kb{(?, r, e)} & {\it what is the \kb{r} of \kb{e} ?}\\
\kb{(?, r, e)} & {\it who is the \kb{r} of \kb{e} ?}\\
\kb{(e, r, ?)} & {\it what is \kb{r} by \kb{e} ?}\\
\kb{(?, r, e)} & {\it who is \kb{e}'s \kb{r} ?}\\
\hline
\end{tabular}
\begin{tabular}{|l|l|}
\hline
\multicolumn{1}{|c|}{{\bf KB Triple}}  & \multicolumn{1}{c|}{{\bf Question Pattern}}\\
\hline
\kb{(?, r, e)} & {\it what is \kb{e}'s \kb{r} ?}\\
\kb{(e, r, ?)} & {\it who is \kb{r} by \kb{e} ?}\\
\kb{(e, r-in, ?)} & {\it when did \kb{e} \kb{r} ?}\\
\kb{(e, r-on, ?)} & {\it when did \kb{e} \kb{r} ?}\\
\kb{(e, r-in, ?)} & {\it when was \kb{e} \kb{r} ?}\\
\kb{(e, r-on, ?)} & {\it when was \kb{e} \kb{r} ?}\\
\kb{(e, r-in, ?)} & {\it where was \kb{e} \kb{r} ?}\\
\kb{(e, r-in, ?)} & {\it where did \kb{e} \kb{r} ?}\\
\hline
\end{tabular}

\end{small}
\end{center}
\vskip -0.1in
\end{table}

\paragraph{Paraphrases}
The automatically generated examples are useful to connect KB triples and natural language.
However, they do not allow for a satisfactory modeling of English language because of their poor wording.
To overcome this issue, we again follow \cite{paralex} and supplement our training data with an indirect supervision signal made of pairs of question paraphrases collected from the \wk website.

On \wk, users can tag pairs of questions as rephrasing of each other.
\cite{paralex} harvested a set of 18M of these question-paraphrase pairs, with 2.4M distinct questions in the corpus. These pairs have been labeled collaboratively. This is cheap but also causes the data to be noisy.
Hence, \cite{paralex} estimated that only 55\% of the pairs were actual paraphrases.
The set of paraphrases is denoted ${\cal P}$ in the following.
By considering all words and tokens appearing in ${\cal P}$ and ${\cal D}$, we end up with a size for the vocabulary ${\cal V}$ of more than 800k.


\section{Embedding-based Model}
\label{sec:model}

Our model ends up learning vector embeddings of symbols, either for  entities or relationships from \rv, or for each word of the vocabulary. 
%


\subsection{Question-KB Triple Scoring}

\paragraph{Architecture}
Our framework concerns the learning of a function $S(q, t)$, based on embeddings, 
that is designed to score the similarity of a question $q$  and a triple $t$ from ${\cal K}$. 

Our scoring approach is inspired by previous work for labeling images with words \cite{wsabie},
which we adapted, replacing images and labels by questions and triples.
Intuitively, it consists of projecting questions, treated as a bag of words (and possibly $n$-grams as well), on the one hand, and triples on the other hand, into a shared embedding space and then computing a similarity measure (the dot product in this paper) between both projections.
The scoring function is then:
$$
S(q,t) = \bbf(q)^\top \bbg(t)
$$
with $\bbf(\cdot)$ a function mapping words from questions into $\Re^k$, $\bbf(q) = \bbV^\top\Phi(q)$. $\bbV$ is the matrix of $\Re^{n_v\times k}$ containing all word embeddings $\bbv$, $\Phi(q)$ is the (sparse)
binary representation of $q$ ($\in \{0,1\}^{n_v}$) indicating absence or presence of words.
Similarly, $\bbg(\cdot)$ is a function mapping entities and relationships from KB triples into $\Re^k$, $\bbg(t) = \bbW^\top\Psi(t)$, $\bbW$ the matrix of $\Re^{n_e\times k}$ containing all entities and relationships embeddings $\bbw$, and $\Psi(q)$ the (sparse)
binary representation of $t$ ($\in \{0,1\}^{n_e}$) indicating absence or presence of entities and relationships.

\ab{Representing questions as a bag of words might seem too limited, however, in our particular setup, syntax is generally simple, and hence quite uninformative.
A question is typically formed by an interrogative pronoun, a reference to a relationship and another one to an entity.
Besides, since lexicons of relationships and entities are rather disjoint, even a bag of words representation should lead to decent performance, up to lexical variability.
There are counter-examples such as {\it What are cats afraid of ?} vs. {\it What are afraid of cats ?} which require different answers, but such cases are rather rare.
Future work could consider adding parse tree features or semantic role labels as input to the embedding model. 
}

Contrary to previous work modeling KBs with embeddings (e.g. \cite{weston-EtAl:2013:EMNLP}), in our model, an entity does not have the same embedding when appearing in the left-hand or in the right-hand side of a triple.
Since, $\bbg(\cdot)$ sums embeddings of all constituents of a triple, we need to use 2 embeddings per entity to encode for the fact that relationships in the KB are not symmetric and so that appearing as a left-hand or right-hand entity is different.

This approach can be easily applied at test time to score any (question, triple) pairs.
Given a question $q$, one can predict the corresponding answer (a triple) $\hat{t}(q)$ with:
\[
    \hat{t}(q)=\arg \max_{t'\in{\cal K}} S(q,t')=\arg \max_{t'\in{\cal K}}  \big(\bbf(q)^\top \bbg(t') \big).
\]

\paragraph{Training by Ranking}

Previous work \cite{wsabie,weston-EtAl:2013:EMNLP} has shown that this kind of model can be 
conveniently trained using a ranking loss.
Hence, given our data set ${\cal D}= \{(q_i, t_i), i=1,\dots,\card{{\cal D}}\}$ consisting of (question, answer triple) training pairs, one could  learn the embeddings using constraints of the form:
\begin{equation*}\label{eq:cons1}
    \forall i,~ \forall t' \neq t_i, ~~ \bbf(q_i)^\top \bbg(t_i)  >    0.1 +  \bbf(q_i)^\top \bbg(t')~~,
\end{equation*}
where $0.1$ is the margin.
That is, we want the triple that labels a given question to be scored higher than other triples in ${\cal K}$ by a margin of $0.1$.
We also enforce a constraint on the norms of the columns of $\bbV$ and $\bbW$, i.e.
 $\forall_i, ||\bbv_i||_2 \leq 1$ and $\forall_j, ||\bbw_j||_2 \leq 1$. 

To train our model, we need positive and negative examples of $(q,t)$ pairs. However, ${\cal D}$ only contains positive samples, for which the triple actually corresponds to the question.
Hence, during training, we use a procedure to corrupt triples. Given $(q,t)\in{\cal D}$, we create a corrupted triple $t'$ with the following method: pick another random triple $t_{tmp}$ from ${\cal K}$, and then, replace with 66\% chance each member of $t$ (left entity, relationship and right entity) by the corresponding element in $t_{tmp}$.
This heuristic creates negative triples $t'$ somewhat similar to their positive counterpart $t$, and is similar to schemes of previous work (e.g. in \cite{collobert:2011b,bordes:12aistats}).

Training the embedding model is carried out by stochastic gradient descent (SGD)
, updating $\bbW$ and $\bbV$ at each step. 
At the start of training the parameters of $\bbf(\cdot)$ and $\bbg(\cdot)$
(the $n_v\times k$ word embeddings in $\bbV$ and the $n_e \times k$ entities and rel. embeddings in $\bbW$) are initialized to random weights (mean 0, standard deviation $\frac{1}{k}$).
Then, we iterate the following steps to train them:
\begin{enumerate}
\item Sample a positive training pair $(q_i, t_i)$ from ${\cal D}$.
\item Create a corrupted triple $t_i'$ ensuring that $t_i'\neq t_i$.
\item Make a stochastic gradient step to minimize $ \big[0.1 - \bbf(q_i)^\top \bbg(t_i) + \bbf(q_i)^\top \bbg(t_i')\big]_+$.
\item Enforce the constraint that each embedding vector is normalized.
\end{enumerate}
The learning rate of SGD is updated during the course of learning using {\sc adagrad} \cite{duchi2011adaptive}. $\big[x\big]_+$ is the positive part of $x$.

\paragraph{Multitask Training with Paraphrases Pairs}
We multitask the training of our model by training on pairs of paraphrases of questions $(q_1, q_2)$ from ${\cal P}$ as well as training on the pseudolabeled data constructed in ${\cal D}$.
We use the same architecture simply replacing $\bbg(\cdot)$ by a copy of $\bbf(\cdot)$. 
This leads to the following function that scores the similarity between two questions:
$$
\Spr{q_1}{q_2} = \bbf(q_1)^\top \bbf(q_2)~.
$$
The matrix $\bbW$ containing embeddings of words is shared between $S$ and $\Sprv$, allowing it to encode information from examples from both ${\cal D}$ and ${\cal P}$.
Training of $\Sprv$ is also conducted with SGD (and {\sc adagrad}) as for $S$, but, in this case, negative examples are created by replacing one of the questions from the pair by another question chosen at random in ${\cal P}$.

During our experiments, $\bbW$ and $\bbV$ were learned by alternating training steps using $S$ and $\Sprv$, switching from one to another at each step.
The initial learning rate was set to $0.1$ and the dimension $k$ of the embedding space to $64$. Training ran for 1 day on a 16 core machine using {\sc hogwild} \cite{recht2011hogwild}.

\subsection{Fine-tuning the Similarity between Embeddings}

The scale of the problem forced us to keep our architecture simple: with $n_e\approx 3.5{\rm M}$ (with 2 embeddings for each entity) and $n_v \approx 800{\rm k}$, we have to learn around $4.3{\rm M}$ embeddings.
With an embedding space of dimension $k=64$, this leads to around $275{\rm M}$ parameters to learn.
The training algorithm must also stay simple to scale on a training set of around $250{\rm M}$ of examples (${\cal D}$ and ${\cal P}$ combined); SGD  appears as the only viable option.

SGD, combined with {\sc adagrad} for adapting the learning rate on the course of training, is a powerful algorithm.
However, the scale of the optimization problem makes it very hard to control and conduct properly until convergence.
When SGD stops after a pre-defined number of epochs, we are almost certain that the problem is not fully solved and that some room for improvement remains: we observed that embeddings were able to often rank correct answers near the top of the candidates list, but not always in the first place.

In this paper, we introduce a way to fine-tune our embedding-based model so that correct answers might end up more often at the top of the list.
Updating the embeddings involves working on too many parameters, but ultimately, these embeddings are meant to be used in a dot-product that computes the similarity between $q$ and $t$.
We propose to learn a matrix $\bbM\in\Re^{k\times k}$ parameterizing the similarity between words and triples embeddings.
The scoring function becomes:
$$
\Sft{q}{t} = \bbf(q)^\top \bbM \bbg(t)~.
$$

$\bbM$ has only $k^2$ parameters and can be efficiently determined by solving the following convex problem (fixing the embedding matrices $\bbW$ and $\bbV$):
\begin{equation*}\label{eq:m}
 {\rm min}_{\bbM}  \frac{\lambda}{2}\norm{\bbM}^2_{F} + \frac{1}{m}\sum_{i=1}^m\big[1 - \Sft{q_i}{t_i} + \Sft{q_i}{t_i'}\big]^2_+~,
\end{equation*}
where $\norm{\bbX}_{F}$ is the Frobenius norm of $\bbX$.
We solve this problem in a few minutes using L-BFGS on a subset of $m=10{\rm M}$ examples from ${\cal D}$. We first use $4{\rm M}$ examples to train and $6{\rm M}$ as validation set to determine the value of the regularization parameter $\lambda$. We then retrain the model on the whole $10{\rm M}$ examples using the selected value, which happened to be $\lambda=1.7\times10^{-5}$.

This fine-tuning is related to learning a new metric in the embedding space, but since the resulting $\bbM$ is not symmetric, it does not define a dot-product.
Still, $\bbM$ is close to a constant factor times identity (as in the original score $S(\cdot)$).
The fine-tuning does not deeply alter the ranking, but, as expected, allows for a slight change in the triples ranking, 
which ends in consistent improvement in performance, as we show in the experiments.

%
%
%








\begin{table}
\caption{Performance of variants of our embedding models and Paralex \cite{paralex} for reranking question-answer pairs from the \wkrv test set.}
\label{tab:res}
\vskip -0.1in
\begin{center}
\begin{tabular}{|l|ccc|c|}
\hline
{\bf Method} & {\bf F1} & {\bf Prec} & {\bf Recall} & {\bf MAP}\\
\hline
Paralex {\it (No. 2-arg)} & 0.40   & {\bf 0.86} & 0.26 & 0.12 \\
Paralex &   0.54 &  0. 77 & 0.42 & 0.22 \\
\hline
Embeddings & 0.68 & 0.68  & 0.68  & 0.37\\
Embeddings {\it (no paraphrase)} & 0.60 & 0.60  & 0.60  & 0.34\\
Embeddings {\it (incl. n-grams)} & 0.68 & 0.68  & 0.68  & 0.39\\
Embeddings+fine-tuning & {\bf 0.73} & 0.73  & {\bf 0.73}  & {\bf 0.42} \\
\hline
\end{tabular}
\end{center}
\end{table}

\section{Experiments}
\label{sec:exp}


\subsection{Evaluation Protocols}

We first detail the data and metrics which were chosen to assess the quality of our embedding model.

\paragraph{Test Set} The data set \wkrv contains no labeled examples but some are needed for evaluating models.
We used the test set which has been created by \cite{paralex} in the following way: (1) they identified 37 questions from a held-out portion of \wk which were likely to have at least one answer in \rv, (2) they added all valid paraphrases of these questions to obtain a set of 691 questions, (3) they ran various versions of their {\sc paralex} system on them to gather candidate triples (for a total of 48k), which they finally hand-labeled.

\begin{figure}[t]
\begin{center}
\centerline{\includegraphics[width=10cm]{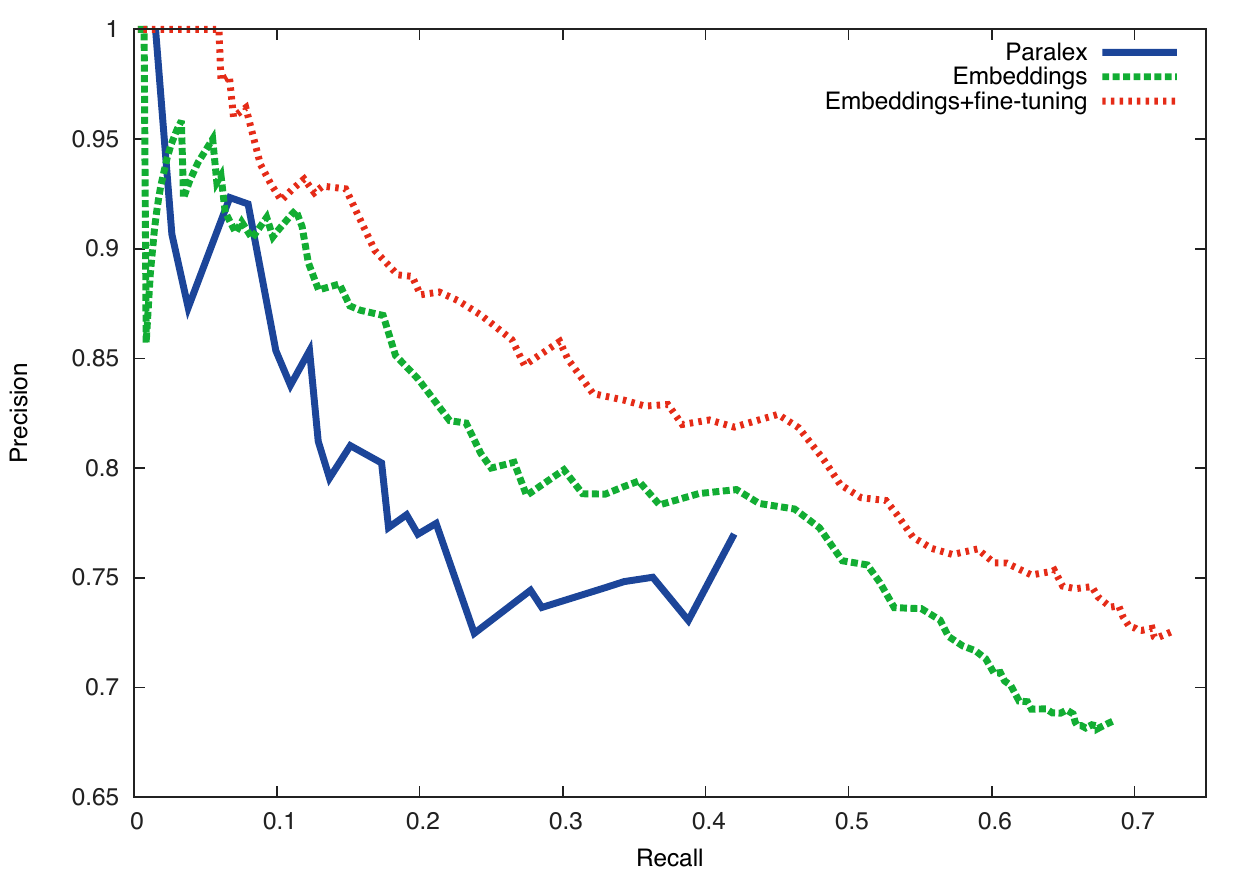}}
\caption{Precision-recall curves of our embedding model and Paralex \cite{paralex} for reranking question-answer pairs from the \wkrv test set.}
\label{fig:pr}
\end{center}
\vskip -0.2in
\end{figure} 

\paragraph{Reranking}

We first evaluated different versions of our model against the {\sc paralex} system in a reranking setting.  
For each question $q$ from the \wkrv test set, we take the provided candidate triples $t$ and rerank them by sorting by the score $S(q,t)$ or $\Sft{q}{t}$ of our model, depending whether we use fine-tuning or not. 
As in \cite{paralex}, we then compute the precision, recall and F1-score of the highest ranked answer as well as the mean average precision (MAP) of the whole output, which measures the average precision over all levels of recall.

\paragraph{Full Ranking}

The reranking setting might be detrimental for {\sc paralex} because our system simply never has to perform a full search for the good answer among the whole \rv KB.
Hence, we also conducted an experiment where, for each of the 691 questions of the \wkrv test set, we ranked all 14M triples from \rv.
We labeled the top-ranked answers ourselves and computed precision, recall and F1-score.\footnote{We provide the top-ranked answers and our labels as supplementary material.}

\begin{table}
\caption{Performance of our embedding model for retrieving answers for questions from the \wkrv test set, among the whole \rv KB (14M candidates).}
\label{tab:resfull}
\vskip -0.1in
\begin{center}
\begin{small}
\begin{tabular}{|l@{}l|c|}
\hline
{\bf Method} && {\bf F1} \\
\hline
Embeddings && 0.16  \\
Embeddings+fine-tuning && 0.22 \\
\hline
Embeddings &+string-matching & 0.48 \\
Embeddings+fine-tuning &+string-matching &  {\bf 0.57} \\
\hline
\end{tabular}
\end{small}
\end{center}
\vskip -0.1in
\end{table}

\subsection{Results}
This section now discusses our empirical performance.

\paragraph{Reranking}

Table~\ref{tab:res} and Figure~\ref{fig:pr} present the results of the reranking experiments.
We compare various versions of our model against two versions of {\sc paralex}, whose results were given in \cite{paralex}.

First, we can see that multitasking with paraphrase data is essential since it improves F1 from 0.60 to 0.68.
Paraphrases allow for the embeddings to encode a richer connection between KB constituents and words, as well as between words themselves.
Note that the \wk data provides word alignment between paraphrases, which we did not use, unlike {\sc paralex}.
We also tried to use n-grams (2.5M most frequent) as well as the words to represent the question, but this
did not bring any improvement, which might at first seem counter-intuitive. We believe this is due to two factors: (1) it is hard to learn good embeddings for n-grams since their frequency is usually very low and (2) our automatically generated questions have a poor syntax and hence, many n-grams in this data set do not make sense.
We actually conducted experiments with several variants of our model, which tried to take the word ordering into account (e.g. with convolutions), and they all failed to outperform our best performance without word order, once again perhaps because the supervision is not clean enough to allow for such elaborated language modeling.
Fine-tuning the embedding model is very beneficial to optimize the top of the list and grants a bump of 5 points of F1: carefully tuning the similarity makes a clear difference.

All versions of our system greatly outperform {\sc paralex}: the fine-tuned model improves the F1-score by almost 20 points and, according to Figure~\ref{fig:pr}, is better in precision for all levels of recall.
{\sc paralex} works by starting with an initial lexicon mapping from the KB to language and then gradually increasing its coverage by iterating  on the \wkrv data.
Most of its predictions come from automatically acquired templates and rules: this allows for a good precision but it is not flexible enough across language variations to grant a satisfying recall. Most of our improvement comes from a much better recall.

However, as we said earlier, this reranking setting is detrimental for {\sc paralex} because {\sc paralex} was evaluated on the task of reranking some of its own predictions.
The results provided for {\sc paralex}, while not corresponding to those of a full ranking among all triples from \rv (it is still reranking among a subset of candidates), concerns an evaluation setting more complicated than for our model.
Hence, we also display the results of a full ranking by our system in the following.

\begin{figure}[t]
\vskip 0.2in
\begin{center}
\centerline{\includegraphics[width=10cm]{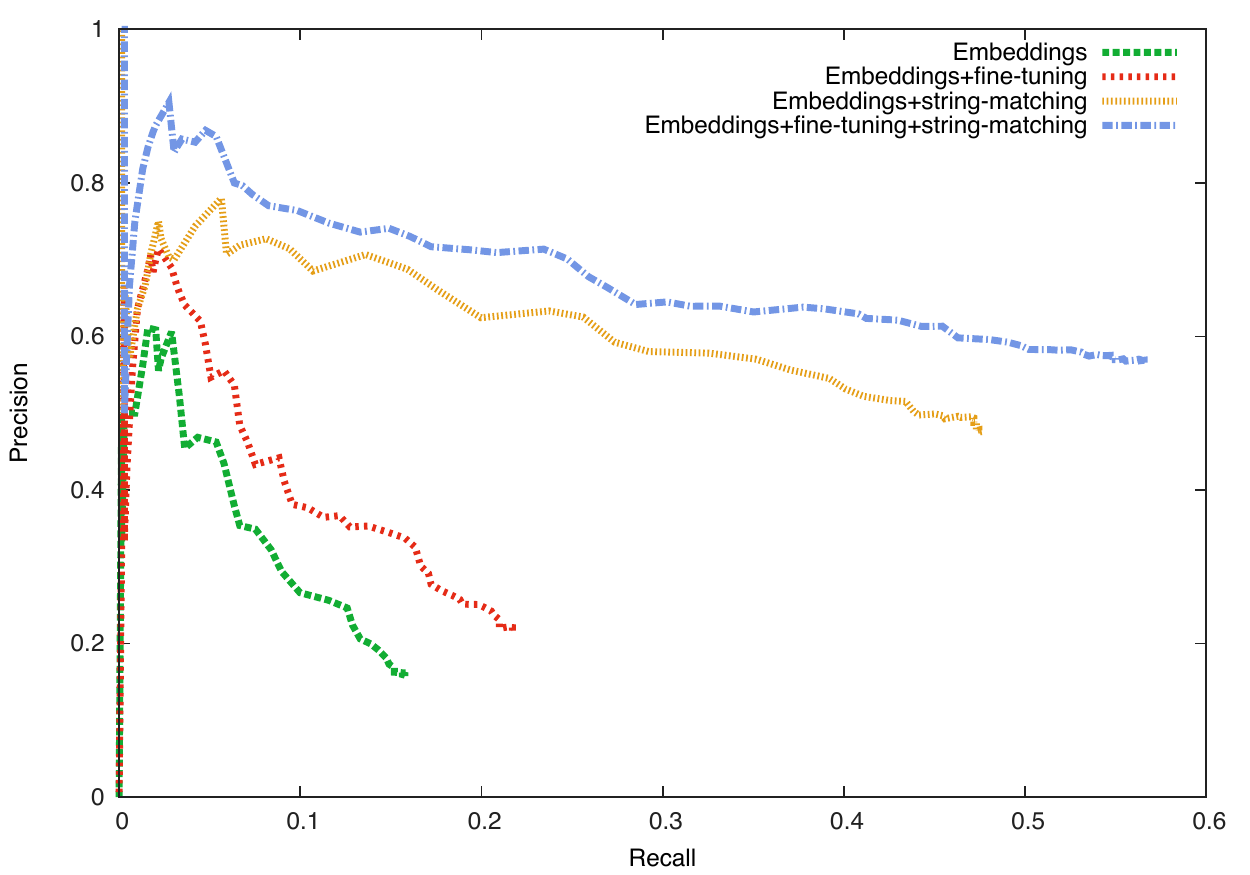}}
\caption{Precision-recall curves for retrieving answers for questions from the \wkrv test set, among the whole \rv KB (14M candidates).}
\label{fig:fpr}
\end{center}
\vskip -0.2in
\end{figure} 

\begin{table*}[t]
\caption{Examples of nearest neighboring entities and relationships from REVERB for some words from our vocabulary. The prefix \kb{L:}, resp. \kb{R:}, indicates the embedding of an entity when appearing in \kb{left-hand} side, resp. \kb{right-hand} side, of triples.}
\label{tab:samples}
\vskip 0.1in
\begin{center}
\resizebox{\linewidth}{!}{
\begin{tabular}{|c|l|}
\hline
\multicolumn{1}{|c|}{{\bf Word}} & \multicolumn{1}{c|}{{\bf Closest entities or relationships from \rv in the embedding space}}\\
\hline
\txt{get rid of}
& 
\kb{get-rid-of.r} \kb{be-get-rid-of.r} \kb{rid-of.r} \kb{can-get-rid-of.r} \kb{will-get-rid-of.r} \kb{should-get-rid-of.r} \\
&
\kb{have-to-get-rid-of.r} \kb{want-to-get-rid-of.r} \kb{will-not-get-rid-of.r} \kb{help-get-rid-of.r} \\
\hline 
\txt{useful}
&
\kb{be-useful-for.r} \kb{be-useful-in.r} \kb{R:wide-range-of-application.e} \kb{can-be-useful-for.r} \\
&
\kb{be-use-extensively-for.r} \kb{be-not-very-useful-for.r} \kb{R:plex-or-technical-algorithm.e} \\
&
\kb{R:internal-and-external-use.e} \kb{R:authoring.e} \kb{R:good-or-bad-purpose.e} \\
\hline 
\txt{radiation}
&
\kb{R:radiation.e} \kb{L:radiation.e} \kb{R:gamma-radiation.e} \kb{L:gamma-radiation.e} \kb{L:x-ray.e} \kb{L:gamma-ray.e} \\
&
\kb{L:cesium-137.e} \kb{R:electromagnetic-radiation.e} \kb{L:external-beam-radiation.e} \kb{L:visible-light.e} \\
\hline 
\txt{barack-obama}
&
\kb{L:president-elect-barack-obama.e} \kb{L:barack-obama.e} \kb{R:barack-obama.e} \kb{L:president-barack-obama.e} \\
&
\kb{L:obama-family.e} \kb{L:sen.-barack-obama.e}\kb{L:president-elect-obama.e} \kb{R:president-barack-obama.e} \\
& 
\kb{L:democratic-presidential-candidate-barack-obama.e} \kb{L:today-barack-obama.e} \\
\hline 
\txt{iphone}
&
\kb{R:iphone.e} \kb{L:iphone.e} \kb{R:t-mobile.e} \kb{R:apple-iphone.e} \kb{L:lot-of-software.e} \kb{L:hotmail.e} \\
&
\kb{R:windows-mobile-phone.e} \kb{L:skype.e} \kb{R:smartphone.e} \kb{R:hd-dvd-player.e} \\
\hline
\end{tabular}
}
\end{center}
\vskip -0.1in
\end{table*}

\paragraph{Full Ranking}

Table~\ref{tab:resfull} and Figure~\ref{fig:fpr} display the results of our model to rank all 14M triples from \rv.
The performance of the plain models is not good (${\rm F1}=0.22$ only for $\Sftv$) because the ranking is degraded by too many candidates. 
But most of these can be discarded beforehand.

We hence decided to filter out some candidates before ranking by using a simple string matching strategy: 
after pos-tagging the question, we construct a set of candidate strings containing (i) all noun phrases that appear less than 1,000 times in \rv, (ii) all proper nouns if any, otherwise the least frequent noun phrase in \rv.
This set of strings is then augmented with the singular form of plural nouns, removing the final "s", if any. Then, only the triples containing at least one of the candidate strings are scored by the model. 
On average, about 10k triples (instead of 14M) are finally ranked for each question, making our approach much more tractable. 

%
%

As expected, string matching greatly improves results, both in precision and recall, and also significantly reduces evaluation time.  

The final F1 obtained by our fine-tuned model is even better then the result of {\sc paralex} in reranking, which is pretty remarkable, because this time, this setting advantages it quite a lot.

\paragraph{Embeddings}

Table~\ref{tab:samples} displays some examples of nearest neighboring entities from \rv for some words from our vocabulary.
As expected, we can see that verbs or adverbs tend to correspond to relationships while nouns refer to entities.
Interestingly, the model learns some synonymy and hyper/hyponymy. For instance,  {\it radiation} is close to \kb{x-ray.e} and {\it iphone} to \kb{smartphone.e}.
This happens thanks to the multitasking with paraphrase data, since in our automatically generated $(q, t)$ pairs, the words {\it radiation} and {\it iphone} are only used for entities with the strings \kb{radiation} and \kb{iphone} respectively in their names.

\begin{table}
\caption{Performance of our embedding model for retrieving answers for questions from the \wq test set, among the whole \rv KB (14M candidates).}
\label{tab:resfull-webq}
\vskip -0.1in
\begin{center}
\begin{small}
\begin{tabular}{|l@{}l|@{\:\:}c@{\:\:}c@{\:\:}|c@{\:\:}|}
\hline
{\bf Method} && {\bf Top-1} & {\bf Top-10} & {\bf F1}\\
\hline
Emb. && 0.025 & 0.094 &  0.025\\
Emb.+fine-tuning && 0.032 & 0.106 & 0.032\\
\hline
Emb. &+string-match. & 0.085 & 0.246& 0.068\\
Emb.+fine-tuning &+string-match. & {\bf 0.094} & {\bf 0.270} & {\bf 0.076} \\
\hline
\end{tabular}
\end{small}
\end{center}
\vskip -0.2in
\end{table}

\subsection{Evaluation on WebQuestions}

Our initial objective was to be able to perform open-domain question answering. 
In this last experimental section, we tend to evaluate how generic our learned system is.
To this end, we propose to ask our model to answer questions coming from another dataset from the literature, but without retraining it with labeled data, just by directly using the parameters learned on \wkrv.

We chose the data set \wq \cite{berant-EtAl:2013:EMNLP}, which consists of natural language questions matched with answers corresponding to entities of \fb: in this case, no triple has to be returned, only a single entity.
We used exact string matching to find the \rv entities corresponding to the \fb answers from the test set of \wq  and obtained 1,538 questions labeled with \rv out of the original 2,034.

Results of different versions of our model are displayed in Table~\ref{tab:resfull-webq}. For each test question, we record the rank of the first \rv triple containing the answer entity. Top-1 and Top-10 are computed on questions for which the system returned at least one answer (around 1,000 questions using string matching), while F1 is computed for all questions.
Of course, performance is not great and can not be directly compared with that of the best system reported in \cite{berant-EtAl:2013:EMNLP} (more than $0.30$ of F1).
\ab{One of the main reasons is that most questions of \wq, such as {\it Who was vice-president after Kennedy died?}, should be represented by multiple triples, a setting for which our system has not been designed.}
Still, for a system trained with almost no manual annotation nor prior information on another dataset, with an other --very noisy-- KB, the results can be seen as particularly promising.
Besides, evaluation is broad since, in \rv, most entities actually appear many times under different names as explained in Section~\ref{sec:qa}.
Hence, there might be higher ranked answers but they are missed by our evaluation script.

\section{Conclusion}

This paper introduces a new framework for learning to perform open question answering with very little supervision.
Using embeddings as its core, our approach can be successfully trained on imperfect labeled data and indirect supervision and significantly outperforms previous work for answering simple factual questions.
Besides, we introduce a new way to fine-tune embedding models for cases where their optimization problem can not be completely solved.

In spite of these promising results, some exciting challenges remain, especially in order to scale up this model to questions with more complex semantics.
Due to the very low supervision signal, our work can only answer satisfactorily simple factual questions, and does not even take into account the word ordering when modeling them.
Further, much more work has to be carried out to encode the semantics of more complex questions into the embedding space.


\bibliography{ecml-arxiv}

\begin{thebibliography}{10}

\bibitem{banko2002askmsr}
M.~Banko, E.~Brill, S.~Dumais, and J.~Lin.
\newblock Askmsr: Question answering using the worldwide web.
\newblock In {\em Proceedings of 2002 AAAI Spring Symposium on Mining Answers
  from Texts and Knowledge Bases}, 2002.

\bibitem{bengio03}
Y.~Bengio, R.~Ducharme, P.~Vincent, and C.~Jauvin.
\newblock A neural probabilistic language model.
\newblock {\em Journal of Machine Learning Research}, 3:1137--1155, 2003.

\bibitem{berant-EtAl:2013:EMNLP}
J.~Berant, A.~Chou, R.~Frostig, and P.~Liang.
\newblock Semantic parsing on {Freebase} from question-answer pairs.
\newblock In {\em Proceedings of the 2013 Conference on Empirical Methods in
  Natural Language Processing}, October 2013.

\bibitem{bollacker2008freebase}
K.~Bollacker, C.~Evans, P.~Paritosh, T.~Sturge, and J.~Taylor.
\newblock Freebase: a collaboratively created graph database for structuring
  human knowledge.
\newblock In {\em Proceedings of the 2008 ACM SIGMOD international conference
  on Management of data}. ACM, 2008.

\bibitem{bordes:12aistats}
A.~Bordes, X.~Glorot, J.~Weston, and Y.~Bengio.
\newblock Joint learning of words and meaning representations for open-text
  semantic parsing.
\newblock In {\em Proc. of the 15th Intern. Conf. on Artif. Intel. and Stat.},
  volume~22. JMLR W\&CP, 2012.

\bibitem{cai-yates:2013:ACL2013}
Q.~Cai and A.~Yates.
\newblock Large-scale semantic parsing via schema matching and lexicon
  extension.
\newblock In {\em Proceedings of the 51st Annual Meeting of the Association for
  Computational Linguistics (Volume 1: Long Papers)}, August 2013.

\bibitem{collobert:2011b}
R.~Collobert, J.~Weston, L.~Bottou, M.~Karlen, K.~Kavukcuoglu, and P.~Kuksa.
\newblock Natural language processing (almost) from scratch.
\newblock {\em Journal of Machine Learning Research}, 12:2493--2537, 2011.

\bibitem{duchi2011adaptive}
J.~Duchi, E.~Hazan, and Y.~Singer.
\newblock Adaptive subgradient methods for online learning and stochastic
  optimization.
\newblock {\em The Journal of Machine Learning Research}, 12, 2011.

\bibitem{ReVerb2011}
A.~Fader, S.~Soderland, and O.~Etzioni.
\newblock Identifying relations for open information extraction.
\newblock In {\em Proceedings of the Conference of Empirical Methods in Natural
  Language Processing ({EMNLP} '11)}, Edinburgh, Scotland, UK, July 27-31 2011.

\bibitem{paralex}
A.~Fader, L.~Zettlemoyer, and O.~Etzioni.
\newblock Paraphrase-driven learning for open question answering.
\newblock In {\em Proceedings of the 51st Annual Meeting of the Association for
  Computational Linguistics}, pages 1608--1618, Sofia, Bulgaria, 2013.

\bibitem{hoffmann2011knowledge}
R.~Hoffmann, C.~Zhang, X.~Ling, L.~Zettlemoyer, and D.~S. Weld.
\newblock Knowledge-based weak supervision for information extraction of
  overlapping relations.
\newblock In {\em Proceedings of the 49th Annual Meeting of the Association for
  Computational Linguistics: Human Language Technologies}, volume~1, 2011.

\bibitem{kwiatkowski-EtAl:2013:EMNLP}
T.~Kwiatkowski, E.~Choi, Y.~Artzi, and L.~Zettlemoyer.
\newblock Scaling semantic parsers with on-the-fly ontology matching.
\newblock In {\em Proceedings of the 2013 Conference on Empirical Methods in
  Natural Language Processing}, October 2013.

\bibitem{kwok2001scaling}
C.~Kwok, O.~Etzioni, and D.~S. Weld.
\newblock Scaling question answering to the web.
\newblock {\em ACM Transactions on Information Systems (TOIS)}, 19(3), 2001.

\bibitem{lao2012reading}
N.~Lao, A.~Subramanya, F.~Pereira, and W.~W. Cohen.
\newblock Reading the web with learned syntactic-semantic inference rules.
\newblock In {\em Proceedings of the 2012 Joint Conference on Empirical Methods
  in Natural Language Processing and Computational Natural Language Learning},
  2012.

\bibitem{dbpedia-swj}
J.~Lehmann, R.~Isele, M.~Jakob, A.~Jentzsch, D.~Kontokostas, P.~N. Mendes,
  S.~Hellmann, M.~Morsey, P.~van Kleef, S.~Auer, and C.~Bizer.
\newblock {DBpedia} - a large-scale, multilingual knowledge base extracted from
  wikipedia.
\newblock {\em Semantic Web Journal}, 2014.

\bibitem{mintz2009distant}
M.~Mintz, S.~Bills, R.~Snow, and D.~Jurafsky.
\newblock Distant supervision for relation extraction without labeled data.
\newblock In {\em Proc. of the Conference of the 47th Annual Meeting of ACL},
  2009.

\bibitem{pomikalek2012building}
J.~Pomik{\'a}lek, M.~Jakub{\'\i}cek, and P.~Rychl{\`y}.
\newblock Building a 70 billion word corpus of english from clueweb.
\newblock In {\em LREC}, pages 502--506, 2012.

\bibitem{recht2011hogwild}
B.~Recht, C.~R{\'e}, S.~J. Wright, and F.~Niu.
\newblock Hogwild!: A lock-free approach to parallelizing stochastic gradient
  descent.
\newblock In {\em Advances in Neural Information Processing Systems (NIPS
  24).}, 2011.

\bibitem{riedel2013relation}
S.~Riedel, L.~Yao, A.~McCallum, and B.~M. Marlin.
\newblock Relation extraction with matrix factorization and universal schemas.
\newblock In {\em Proceedings of NAACL-HLT}, 2013.

\bibitem{socher2013recursive}
R.~Socher, A.~Perelygin, J.~Y. Wu, J.~Chuang, C.~D. Manning, A.~Y. Ng, and
  C.~Potts.
\newblock Recursive deep models for semantic compositionality over a sentiment
  treebank.
\newblock In {\em Proceedings of the Conference on Empirical Methods in Natural
  Language Processing (EMNLP)}, 2013.

\bibitem{unger2012template}
C.~Unger, L.~B{\"u}hmann, J.~Lehmann, A.-C. Ngonga~Ngomo, D.~Gerber, and
  P.~Cimiano.
\newblock Template-based question answering over rdf data.
\newblock In {\em Proceedings of the 21st international conference on World
  Wide Web}, 2012.

\bibitem{voorhees2000building}
E.~M. Voorhees and D.~M. Tice.
\newblock Building a question answering test collection.
\newblock In {\em Proceedings of the 23rd annual international ACM SIGIR
  conference on Research and development in information retrieval}. ACM, 2000.

\bibitem{wsabie}
J.~Weston, S.~Bengio, and N.~Usunier.
\newblock Large scale image annotation: learning to rank with joint word-image
  embeddings.
\newblock {\em Machine learning}, 81(1), 2010.

\bibitem{weston-EtAl:2013:EMNLP}
J.~Weston, A.~Bordes, O.~Yakhnenko, and N.~Usunier.
\newblock Connecting language and knowledge bases with embedding models for
  relation extraction.
\newblock In {\em Proceedings of the 2013 Conference on Empirical Methods in
  Natural Language Processing}, pages 1366--1371, 2013.

\bibitem{yahya2012natural}
M.~Yahya, K.~Berberich, S.~Elbassuoni, M.~Ramanath, V.~Tresp, and G.~Weikum.
\newblock Natural language questions for the web of data.
\newblock In {\em Proceedings of the Conference on Empirical Methods in Natural
  Language Processing}, 2012.

\end{thebibliography}
\bibliographystyle{abbrv}

\end{document}